\PassOptionsToPackage{bookmarksnumbered,unicode}{hyperref} 
\PassOptionsToPackage{hyphens}{url}           
\RequirePackage{hyperref}                     

\documentclass[sigconf]{acmart}

\renewcommand\footnotetextcopyrightpermission[1]{} 
\AtBeginDocument{%
  }

\setcopyright{acmcopyright}
\copyrightyear{2018}
\acmYear{2018}
\acmDOI{XXXXXXX.XXXXXXX}

\acmConference[ICIIT]{2026 International Conference on Intelligent Information Technology}{March 05--08,
  2026}{Da Nang, Vietnam}
\acmPrice{15.00}
\acmISBN{978-1-4503-XXXX-X/18/06}

\usepackage{algorithm}
\usepackage{algpseudocode}
\usepackage{subcaption}

\begin{document}

\title{Bi-Layer Ant Colony Optimization for Multi-Robot Task Allocation and Routing in Delivery Applications}

\author{Le Na Nguyen}
\affiliation{%
  \institution{Fulbright University Vietnam}
  \city{Ho Chi Minh City}
  \country{Vietnam}
}

\author{Thanh Long Nguyen}
\affiliation{%
  \institution{Fulbright University Vietnam}
  \city{Ho Chi Minh City}
  \country{Vietnam}
}

\author{Thanh Thao Ton Nu}
\affiliation{%
  \institution{Fulbright University Vietnam}
  \city{Ho Chi Minh City}
  \country{Vietnam}
}

\author{Quan Le}
\affiliation{%
  \institution{Fulbright University Vietnam}
  \city{Ho Chi Minh City}
  \country{Vietnam}
}
\email{quan.le@fulbright.edu.vn}

\author{Manh Duong Phung}
\affiliation{%
  \institution{College of Engineering and Computer Science, VinUniversity}
  \city{Hanoi}
  \country{Vietnam}
}
\email{duong.pm@vinuni.edu.vn}

\renewcommand{\shortauthors}{Le Na et al.}

\begin{abstract}
This paper addresses the multi-robot task allocation (MRTA) problem, which is essential for delivery and logistics applications. Our approach first defines a new cost function that transforms the MRTA into a unified optimization problem capturing both task assignment and routing. A bi-layer ant colony optimization (ACO) algorithm is then introduced, integrating two interdependent decision layers within a single colony process to solve the problem. This hierarchical framework enables simultaneous optimization of task allocation and route planning across multiple robots. Comparative experiments with mixed-integer linear programming (MILP) and particle swarm optimization (PSO) demonstrate that the proposed bi-layer ACO achieves the shortest total travel distance and fastest completion time across all task sizes. Specifically, it reduces total travel distance by up to 17.7\% and completion time by nearly 20\% compared with baseline methods. These results confirm the efficiency, scalability, and reliability of the proposed bi-layer ACO for multi-robot delivery tasks.
\end{abstract}

\maketitle
\pagestyle{empty}
\thispagestyle{empty}

\section{Introduction}
Task allocation plays a key role in coordinating fleets of mobile robots for pickup, transport, and delivery operations in logistics and service applications. It involves assigning specific delivery tasks to individual robots and determining their execution sequence to achieve collective objectives. Serving as the decision-making core, task allocation links robot capabilities, task requirements, and environmental conditions. An effective task allocation strategy enhances the efficiency, scalability, and reliability of multi-robot delivery systems by reducing redundant motion, minimizing total travel distance, and improving overall delivery speed and service quality.

Methods for multi-robot task allocation (MRTA) have evolved from exact optimization formulations to distributed, metaheuristic, and hybrid strategies that balance optimality and scalability. Popular approaches formulated MRTA as assignment or routing problems and solve them using mixed-integer linear programming (MILP) or branch-and-bound search \cite{lippi2021mixed, martin2021multi}. These methods are effective for small to medium-scale settings where precision is essential. To address dynamic and uncertain environments, distributed and market-based architectures such as the consensus-based bundle algorithm (CBBA) are introduced to allocate tasks through iterative bidding and consensus among agents \cite{Huo2024, mahato2023consensus, zitouni2020distributed}. These algorithms achieve near-optimal performance while reducing computation and communication costs. Hybrid extensions such as CBBA combined with ant colony system further improve scalability and convergence by integrating heuristic exploration with distributed coordination \cite{zitouni2020distributed}. Similarly, hierarchical or group-based auctions have been introduced for large-scale domains, where task clustering is used to balance workloads \cite{xue2019task, bai2022group}.

Recent research has increasingly adopted metaheuristic and learning-based MRTA frameworks to address high-dimensional and dynamic task environments. Particle swarm optimization (PSO) and its multi-objective variants (MOPSO) have been applied to cooperative MRTA to minimize total team travel distance and balance individual workloads \cite{wei2020particle}. Other evolutionary algorithms, including genetic algorithms and ant colony optimization (ACO), have been used for large-scale routing and scheduling problems to balance computational efficiency and solution quality \cite{9237316, wang2022cooperative}. In parallel, reinforcement learning (RL) and deep neural architectures have emerged as adaptive alternatives that learn task allocation policies under uncertainty and partial observability \cite{miao2024multi, devin2017learning}. Other approaches combine optimization-based initialization with learning-driven refinement to enhance adaptability and robustness. These studies highlight ongoing progress toward hybrid MRTA systems that combine optimization and learning for improved adaptability. However, most existing approaches still address task assignment, task sequencing, and execution as independent stages that are optimized separately. This separation prevents global coordination and often leads to suboptimal routes and redundant motion. The lack of a unified algorithm that optimizes both task assignment and routing across multiple robots remains a key limitation,  restricting overall system efficiency and optimality in real-world delivery and logistics scenarios.

To overcome this limitation, this work introduces a bi-layer ACO framework that simultaneously optimizes task assignment and task sequencing for multiple robots. A key contribution of this study is the formulation of a new objective function with assignment and sequencing constraints, which transforms the MRTA into a coupled optimization problem. This formulation captures the interdependence between global task distribution and local route planning so that both are optimized together rather than in separate stages. On this basis, the proposed bi-layer ACO employs two interconnected pheromone matrices and heuristic components to jointly learn optimal task assignments and route sequences. This unified design reduces redundant travel and improves overall mission efficiency in complex multi-robot delivery tasks.

\section{Problem Formulation}

This work addresses the MRTA problem for a team of mobile robots executing delivery tasks in a shared environment. Each robot is required to visit a set of assigned pickup and drop-off locations so that all tasks are completed while minimizing the total travel distance of the entire fleet. The optimization jointly determines (i) which robot performs each task and (ii) the order in which the assigned tasks are executed.

Let $\mathcal{R} = \{1, \dots, n\}$ denote the set of robots and $\mathcal{T} = \{1, \dots, m\}$ the set of tasks.  
Each task $j \in \mathcal{T}$ consists of a pickup position $\mathbf{p}_j$ and a drop-off position $\mathbf{d}_j$.  
Each robot $i \in \mathcal{R}$ starts from an initial position $\mathbf{s}_i$ and travels along a route that begins at its start location, visits its assigned pickups and drop-offs in sequence, and ends after its final delivery.

A task must be executed exactly once by a single robot.  
Let $x_{i,j}$ be the binary assignment variable:
\begin{equation}
x_{i,j} =
\begin{cases}
1, & \text{if task } j \text{ is assigned to robot } i,\\
0, & \text{otherwise.}
\end{cases}
\label{eq:binary_constraint}
\end{equation}

The order in which each robot performs its assigned tasks is represented by a sequence variable $\pi_i = (\pi_i(1), \pi_i(2), \dots)$,  
which is a permutation of the task indices assigned to robot $i$.  
The route of robot $i$ is expressed as
\begin{equation}
\mathbf{s}_i 
\rightarrow 
\mathbf{p}_{\pi_i(1)} \rightarrow \mathbf{d}_{\pi_i(1)} 
\rightarrow 
\mathbf{p}_{\pi_i(2)} \rightarrow \mathbf{d}_{\pi_i(2)} 
\rightarrow \cdots .
\label{eq:route_definition}
\end{equation}

For instance, if robot $i$ is assigned tasks $\{2,4,7\}$ with sequence $\pi_i = (4,2,7)$, then its trajectory becomes:
\begin{equation}
\mathbf{s}_i \rightarrow 
\mathbf{p}_4 \rightarrow \mathbf{d}_4 \rightarrow 
\mathbf{p}_2 \rightarrow \mathbf{d}_2 \rightarrow 
\mathbf{p}_7 \rightarrow \mathbf{d}_7.
\end{equation}

\subsection{Objective Function}

The optimization aims to minimize the total travel distance of all robots:
\begin{equation}
\min_{x, \pi} 
\; C = 
\sum_{i \in \mathcal{R}} D_i(x, \pi_i),
\label{eq:objective}
\end{equation}
where $D_i(x, \pi_i)$ is the total travel distance of robot $i$ defined as
\begin{equation}
D_i(x, \pi_i) =
d(\mathbf{s}_i, \mathbf{p}_{\pi_i(1)}) +
\sum_{k=1}^{|\pi_i|}
d(\mathbf{p}_{\pi_i(k)}, \mathbf{d}_{\pi_i(k)}) +
\sum_{k=1}^{|\pi_i|-1}
d(\mathbf{d}_{\pi_i(k)}, \mathbf{p}_{\pi_i(k+1)}),
\label{eq:distance}
\end{equation}
where $d(\cdot)$ denotes the Euclidean distance between two locations.  
This cost structure captures all motion segments in the delivery process: 
\textit{start-to-pickup}, \textit{pickup-to-drop-off}, and \textit{drop-off-to-next-pickup}.

\subsection{Constraints}
Each task must be assigned to exactly one robot:
\begin{equation}
\sum_{i \in \mathcal{R}} x_{i,j} = 1,
\quad \forall j \in \mathcal{T}.
\label{eq:assignment_constraint}
\end{equation}
Besides, each robot’s route must form a valid sequence of assigned tasks, and each task can appear at most once:
\begin{equation}
\pi_i \text{ is a valid permutation of } 
\{ j \in \mathcal{T} \mid x_{i,j}=1 \}.
\label{eq:permutation_constraint}
\end{equation}

\subsection{Optimization Problem}

The complete MRTA formulation is therefore:
\begin{align}
\min_{x, \pi} \quad & 
\sum_{i \in \mathcal{R}} D_i(x, \pi_i), 
\label{eq:final_objective}\\[4pt]
\text{s.t.} \quad &
\sum_{i \in \mathcal{R}} x_{i,j} = 1, \quad \forall j \in \mathcal{T}, \nonumber\\
& x_{i,j} \in \{0,1\}, \quad \forall i \in \mathcal{R}, \, j \in \mathcal{T}, \nonumber\\
& \pi_i \text{ is a valid permutation of assigned tasks}, \quad \forall i \in \mathcal{R}. \nonumber
\end{align}

This formulation captures the coupled nature of the MRTA problem in which both assignment and sequencing decisions affect the overall cost.  

\section{Methodology}
To address the optimization problem defined in (\ref{eq:final_objective}), we propose a bi-layer ACO algorithm that jointly optimizes task assignment and task sequencing for multiple robots. The following subsections outline the fundamental principles of ACO and its adaptation to solve the MRTA problem.

\subsection{Overview of Ant Colony Optimization}
ACO is a population-based metaheuristic inspired by the foraging behavior of real ants. Ants collectively find the shortest paths between their nest and food sources by depositing pheromones
along the paths they traverse. The higher the pheromone concentration on a particular path, the more likely it is to be selected by subsequent ants, allowing the colony to gradually reinforce efficient routes. 
Mathematically, each ant constructs a solution incrementally according to a stochastic decision rule that balances pheromone intensity and heuristic information:
\begin{equation}
P_{i,j} = 
\frac{(\tau_{i,j})^{\alpha} (\eta_{i,j})^{\beta}}
{\sum_{r} (\tau_{r,j})^{\alpha} (\eta_{r,j})^{\beta}},
\label{eq:aco_transition}
\end{equation}
where $\tau_{i,j}$ represents the pheromone level, $\eta_{i,j}$ is the heuristic desirability, and $\alpha$, $\beta$ control their relative influence. 
After each iteration, pheromone trails are updated based on solution quality:

\begin{equation}
\tau_{i,j} \leftarrow (1 - \rho)\,\tau_{i,j}
+ \sum_{a \in \mathcal{E}} \frac{Q}{C^{(a)}}\,
\mathbf{1}\!\big[(i,j) \in \text{solution of ant } a\big],
\label{eq:aco_general_update_combined}
\end{equation}
where $\rho \in (0,1)$ is the evaporation rate, $Q$ is the pheromone deposit factor, $\mathcal{E}$ is the set of depositing ants, $C^{(a)}$ is the cost of the solution constructed by ant~$a$, and $\mathbf{1}[\cdot]$ is an indicator function equal to~1 if the corresponding component appears in that ant’s solution. This mechanism iteratively reinforces good solutions while maintaining exploration capability.

\subsection{Bi-Layer ACO for the MRTA Problem}
We adapt the classical ACO algorithm to solve the MRTA problem, where each ant constructs a complete solution by jointly optimizing two interdependent decision layers:

\begin{enumerate}
    \item The \textit{task assignment layer}, which determines the allocation of tasks to robots, i.e., $x_{i,j}$ indicating whether robot $i$ performs task $j$.
    \item The \textit{task sequencing layer}, which determines the optimal visiting order $\pi_i$ of tasks assigned to each robot $i$.
\end{enumerate}

Unlike conventional hierarchical approaches where these layers are solved separately, the bi-layer ACO integrates them into a single colony process.  
Each ant builds both assignment and sequencing decisions in one construction phase, guided by two distinct pheromone matrices and corresponding heuristic components.  

\textbf{Pheromone and heuristic representation:}
 In bi-layer ACO, two pheromone matrices and their corresponding heuristics are maintained:
\begin{itemize}
    \item $\boldsymbol{\tau}^{\text{assign}} \in \mathbb{R}^{m_{\text{tasks}} \times n_{\text{robots}}}$ represents the learned desirability of assigning task $j$ to robot $i$.
    \item $\boldsymbol{\tau}^{\text{seq}} \in \mathbb{R}^{m_{\text{tasks}} \times m_{\text{tasks}}}$ represents the desirability of transitioning from task $j$ to task $k$ within the same robot’s sequence.
\end{itemize}

Heuristic information encourages shorter travel distances:
\begin{equation}
\eta^{\text{assign}}_{i,j}=\frac{1}{d(\mathbf{s}_i,\mathbf{p}_j)+\epsilon},
\qquad
\eta^{\text{seq}}_{j,k}=\frac{1}{d(\mathbf{d}_j,\mathbf{p}_k)+\epsilon},
\label{eq:heuristic}
\end{equation}
where $\epsilon$ is a small constant to prevent division by zero.

\textbf{Decision layers:}
At the \textit{assignment layer}, each task $j$ is probabilistically assigned to a robot $i$ according to:
\begin{equation}
P(i \mid j)=
\frac{(\tau^{\text{assign}}_{i,j})^{\alpha}(\eta^{\text{assign}}_{i,j})^{\beta}}
{\sum_{r\in\mathcal{R}}(\tau^{\text{assign}}_{r,j})^{\alpha}(\eta^{\text{assign}}_{r,j})^{\beta}}.
\label{eq:probability_assign}
\end{equation}

At the \textit{sequencing layer}, once a robot’s task set is known, the sequence is built by first selecting the task whose pickup is closest to the robot’s start:
\begin{equation}
\pi_i(1)=\arg\min_{j\in\mathcal{U}_i} d(\mathbf{s}_i,\mathbf{p}_j).
\label{eq:seq_start}
\end{equation}
The next task $k$ is then sampled  based on pheromone and heuristic information of task transitions:
\begin{equation}
P(k\mid j,i)=
\frac{(\tau^{\text{seq}}_{j,k})^{\alpha}(\eta^{\text{seq}}_{j,k})^{\beta}}
{\sum_{u\in\mathcal{U}_i}(\tau^{\text{seq}}_{j,u})^{\alpha}(\eta^{\text{seq}}_{j,u})^{\beta}},
\label{eq:probability_seq}
\end{equation}
where $\mathcal{U}_i$ denotes the set of unvisited tasks assigned to robot $i$.  

\textbf{Pheromone update:}  
After all ants have constructed their complete MRTA solutions, pheromone trails on both layers are updated through a combination of evaporation and reinforcement. The update equations for the two pheromone layers are defined as:
\begin{equation}
\begin{aligned}
\tau^{\text{assign}}_{i,j} &\leftarrow (1-\rho)\,\tau^{\text{assign}}_{i,j} + \sum_{a \in \mathcal{E}}\frac{Q}{C^{(a)}}\,
\mathbf{1}\!\big[(i,j) \in \text{assignment of ant } a\big],
\end{aligned}
\label{eq:pher_update_assign}
\end{equation}
\begin{equation}
\begin{aligned}
\tau^{\text{seq}}_{j,k} &\leftarrow (1-\rho)\,\tau^{\text{seq}}_{j,k} \\
&+ \sum_{a \in \mathcal{E}}\frac{Q}{C^{(a)}}
\sum_{i\in\mathcal{R}}\sum_{\ell=1}^{|\pi^{(a)}_i|-1}
\mathbf{1}\!\big[(j,k) \in \text{sequence of robot } i \text{ in ant } a\big],
\end{aligned}
\label{eq:pher_update_seq}
\end{equation}

Equation~(\ref{eq:pher_update_assign}) updates the pheromone levels on the task-robot matrix, where each element $\tau^{\text{assign}}_{i,j}$ represents the learned desirability of assigning task $j$ to robot $i$.  
The first term, $(1-\rho)\,\tau^{\text{assign}}_{i,j}$, models pheromone evaporation to gradually reduce the influence of outdated task-robot assignments. The second term deposits new pheromone proportional to $\frac{Q}{C^{(a)}}$ for each ant $a$ that includes the pair $(i,j)$ in its solution.  

Equation~(\ref{eq:pher_update_seq}) governs pheromone updates on the sequencing matrix, where each entry $\tau^{\text{seq}}_{j,k}$ indicates the desirability of performing task $k$ immediately after task $j$.  
The first term again represents evaporation, whereas the second term accumulates pheromone deposits from all ants and all robots within each ant’s solution.  
The outer summation over ants, $\sum_{a \in \mathcal{E}}$, accounts for contributions from every depositing ant, each weighted by $\frac{Q}{C^{(a)}}$ so that better solutions exert stronger influence.  
The inner summations, $\sum_{i\in\mathcal{R}}\sum_{\ell=1}^{|\pi^{(a)}_i|-1}$, iterate through each robot~$i$ and each consecutive task pair $(\pi^{(a)}_i(\ell), \pi^{(a)}_i(\ell+1))$ in its route to ensure that pheromone is reinforced precisely on transitions $(j,k)$ appearing in the ant’s sequence.  

Through this layered pheromone update mechanism, the colony collectively reinforces promising task assignments and routing patterns to reach an optimized solution that minimizes the total travel cost across all robots.

\begin{algorithm}[htbp]
\caption{Bi-Layer ACO for MRTA}
\label{alg:BiLayer-ACO-MRTA}
\begin{algorithmic}[1]
\Require 
Number of robots $n$, tasks $m$, number of ants $N_{\text{ants}}$, 
iterations $N_{\text{iter}}$, parameters $\alpha$, $\beta$, $\rho$, $Q$
\Ensure 
Best assignment $\{x_{i,j}^*\}$, sequences $\{\pi_i^*\}$, and minimum total cost $C^*$

\State Initialize pheromone matrices 
$\boldsymbol{\tau}^{\text{assign}}$ and $\boldsymbol{\tau}^{\text{seq}}$ 
\State Compute heuristic matrices $\boldsymbol{\eta}^{\text{assign}}$ and $\boldsymbol{\eta}^{\text{seq}}$ using (\ref{eq:heuristic})

\For{$t = 1$ to $N_{\text{iter}}$}
    \For{each ant $a = 1$ to $N_{\text{ants}}$}
        \State Initialize empty assignment $x_{i,j}^{(a)}$ and task lists $\{\mathcal{U}_i^{(a)}\}$
        
        \State \textbf{// Assignment Layer:}
        \For{each task $j = 1, \dots, m$}
            \State Select a robot $i$ probabilistically using (\ref{eq:probability_assign})
            \State Assign task $j$ to robot $i$: $x_{i,j}^{(a)} \leftarrow 1$; add $j$ to $\mathcal{U}_i^{(a)}$
        \EndFor

        \State \textbf{// Sequencing Layer:}
        \For{each robot $i = 1, \dots, n$}
            \If{$|\mathcal{U}_i^{(a)}| > 0$}
                \State Initialize first task using (\ref{eq:seq_start})
                \While{there are unvisited tasks in $\mathcal{U}_i^{(a)}$}
                    \State Select next task $k$ using (\ref{eq:probability_seq})
                    \State Append $k$ to $\pi_i^{(a)}$ and remove it from $\mathcal{U}_i^{(a)}$
                \EndWhile
            \EndIf
        \EndFor

        \State \textbf{// Evaluation:}
        \State Compute total cost $C^{(a)}$ using (\ref{eq:objective})
        \If{$C^{(a)} < C^*$}
            \State Update best solution: $C^* \leftarrow C^{(a)}$, 
            $x_{i,j}^* \leftarrow x_{i,j}^{(a)}$, 
            $\pi_i^* \leftarrow \pi_i^{(a)}$
        \EndIf
    \EndFor

    \State \textbf{// Pheromone Update:}
    \State Update $\boldsymbol{\tau}^{\text{assign}}$ using (\ref{eq:pher_update_assign})
    \State Update $\boldsymbol{\tau}^{\text{seq}}$ using (\ref{eq:pher_update_seq})
\EndFor

\State \Return $\{x_{i,j}^*\}$, $\{\pi_i^*\}$, $C^*$
\end{algorithmic}
\end{algorithm}

\subsection{Bi-Layer ACO Implementation}
Algorithm~\ref{alg:BiLayer-ACO-MRTA} presents the implementation of the proposed bi-layer ACO algorithm. 
It begins with the initialization of the pheromone matrices $\boldsymbol{\tau}^{\text{assign}}$ and $\boldsymbol{\tau}^{\text{seq}}$ 
along with their corresponding heuristic matrices $\boldsymbol{\eta}^{\text{assign}}$ and $\boldsymbol{\eta}^{\text{seq}}$ as defined in~(\ref{eq:heuristic}).  
The algorithm then iterates over $N_{\text{iter}}$ iterations. 
In each iteration, $N_{\text{ants}}$ ants construct candidate MRTA solutions through the following stages:

\begin{enumerate}
    \item \textbf{Task assignment layer:}  
    Each task $j$ is probabilistically assigned to a robot $i$ according to~(\ref{eq:probability_assign}), 
    forming the assignment matrix $x_{i,j}$ and the corresponding task set for each robot.
    
    \item \textbf{Task sequencing layer:}  
    For each robot $i$, a task sequence $\pi_i$ is generated.  
    The first task is selected using~(\ref{eq:seq_start}), while the remaining tasks are chosen sequentially based on~(\ref{eq:probability_seq}).  
    
    \item \textbf{Cost evaluation:}  
    The total cost $C$ of the constructed solution is computed as the sum of all robot travel distances in~(\ref{eq:objective}).  
    The best solution is updated whenever a lower-cost configuration is identified. 
    
    \item \textbf{Pheromone update:}  
    After all ants complete their routes, pheromone trails on both layers are updated according to~(\ref{eq:pher_update_assign}) and~(\ref{eq:pher_update_seq}).
\end{enumerate}

This iterative process guides subsequent ants toward task allocations and routing patterns that produce shorter total distances and faster completion times.  
The final output includes:  
(i) the optimal or near-optimal task-to-robot assignment $\{x_{i,j}^*\}$,  
(ii) the optimized task sequence $\{\pi_i^*\}$ for each robot, and  
(iii) the corresponding minimum total cost $C^*$.  

\section{Results}
This section presents the evaluation results of the proposed bi-layer ACO algorithm for multi-robot task allocation and sequencing. A series of simulation experiments were conducted to assess its performance in terms of total travel distance and task completion time.

\subsection{Evaluation Setup}

\begin{figure}[t]
    \centering
    \begin{subfigure}[b]{0.8\columnwidth}
        \centering
        \includegraphics[width=\textwidth]{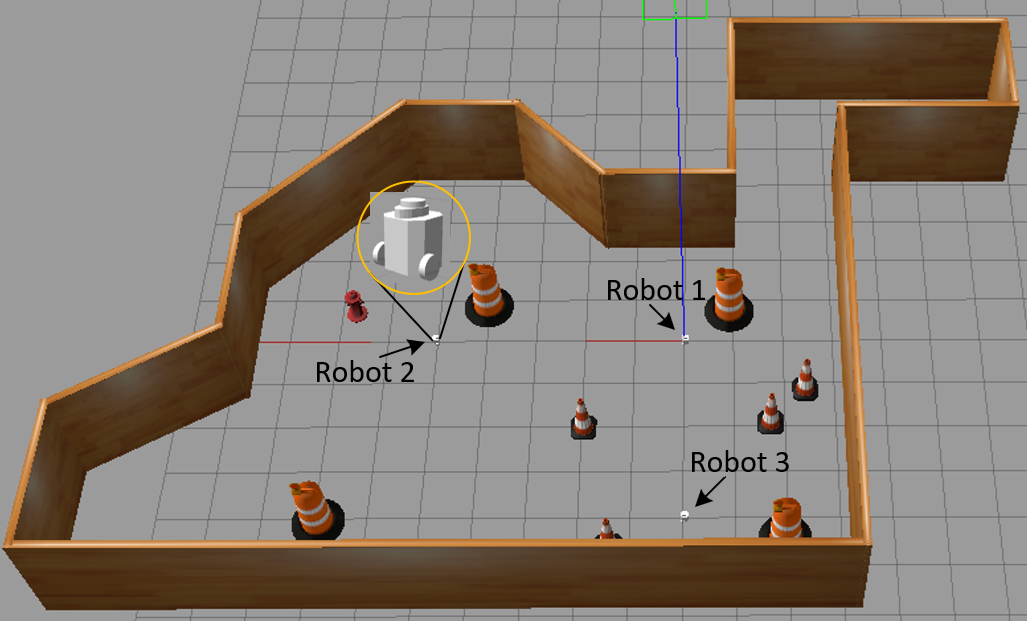}
        \caption{Evaluation scenario with three robots.}
        \label{fig:robot}
    \end{subfigure}
    \hfill
    \begin{subfigure}[b]{0.8\columnwidth}
        \centering
        \includegraphics[width=\textwidth]{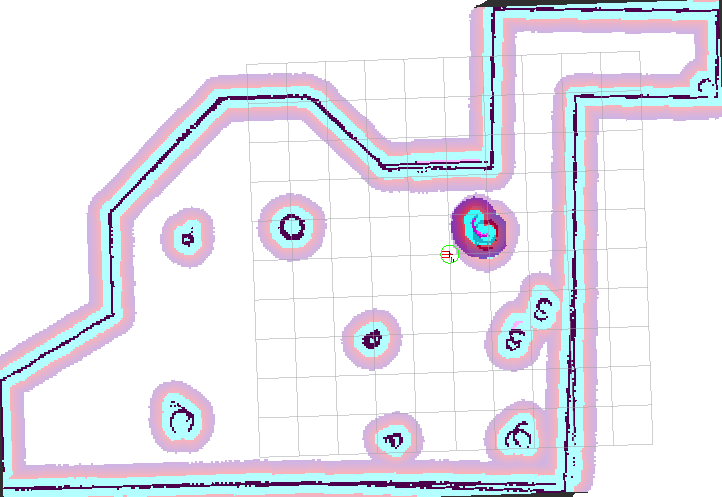}
        \caption{Environment costmap of Robot~1.}
        \label{fig:map}
    \end{subfigure}
    \caption{Simulation environment and robot setup.}
    \label{fig:setup}
\end{figure}

The evaluation is conducted in ROS~2 Humble with Gazebo, a standard simulation platform for robot navigation. 
As illustrated in Figure~\ref{fig:robot}, the simulated environment has a size of $10\,\text{m} \times 15\,\text{m}$ and contains several static obstacles. 
Three homogeneous robots are employed, each modeled after the TurtleBot3 equipped with a 2D LiDAR sensor. 
Figure~\ref{fig:map} presents the costmap of the environment obtained from Robot~1’s LiDAR data.

In all experiments, the initial robot positions are fixed at $(0,0)$, $(5,0)$, and $(0,4)$. 
Three scenarios with $5$, $10$, and $20$ tasks are evaluated, where pickup and drop-off locations are randomly generated at runtime. 
The proposed bi-layer ACO algorithm is first executed to assign tasks and determine the visiting sequence for each robot. 
The output is an ordered list of waypoints representing pickup and drop-off locations that each robot must visit. 
A modified \texttt{Nav2} ROS~2 package is then used to navigate the robots through their assigned waypoints autonomously.

For performance benchmarking, the proposed method is compared with two representative MRTA baselines: 
Mixed Integer Linear Programming (MILP) \cite{7393463,floudas2005mixed} and 
Particle Swarm Optimization (PSO) \cite{kennedy1995particle,ai2009particle}.

\subsection{Evaluation Results}
Table \ref{tab:agg_summary} summarizes the average total travel distance and total completion time obtained by ACO, PSO, and MILP across 10 runs for each algorithm and task size. For all task levels, ACO achieves the lowest mean distance and time, demonstrating superior optimization efficiency and scalability. At 5 tasks, ACO reduces total distance by approximately 17.7\% and 9.8\% compared to MILP and PSO, respectively, while also completing tasks faster. As the task number increases to 10 and 20, all algorithms show higher distances and times due to greater problem complexity. However, ACO consistently maintains the smallest mean and comparable or lower variance, indicating both effective convergence and solution stability.

\begin{table}[h!]
\centering
\small
\caption{Comparison results (mean ± std) for total distance and total time.}
\label{tab:agg_summary}
\begin{tabular}{lcccc}
\toprule
\textbf{Tasks} & \textbf{Algorithm} & \textbf{Distance (m)} & \textbf{Time (s)} \\
\midrule
5  & ACO  & $\textbf{20.00} \pm \textbf{3.43}$ & $\textbf{132.95} \pm \textbf{20.53}$ \\
   & PSO  & $22.17 \pm 3.37$ & $138.66 \pm 19.37$ \\
   & MILP & $24.30 \pm 2.69$ & $166.88 \pm 36.04$ \\
\midrule
10 & ACO  & $\textbf{41.10} \pm \textbf{8.10}$ & $\textbf{287.62} \pm \textbf{38.78}$ \\
   & PSO  & $43.19 \pm 8.24$ & $300.85 \pm 21.29$ \\
   & MILP & $44.19 \pm 8.24$ & $319.43 \pm 74.03$ \\
\midrule
20 & ACO  & $\textbf{90.58} \pm \textbf{12.06}$ & $\textbf{653.48} \pm \textbf{64.20}$ \\
   & PSO  & $101.19 \pm 11.67$ & $715.93 \pm 112.97$ \\
   & MILP & $98.44 \pm 10.83$ & $685.09 \pm 123.80$ \\
\bottomrule
\end{tabular}
\end{table}

\begin{figure*}[t]
    \centering
    \begin{subfigure}[b]{0.32\textwidth}
        \centering
        \includegraphics[width=\linewidth]{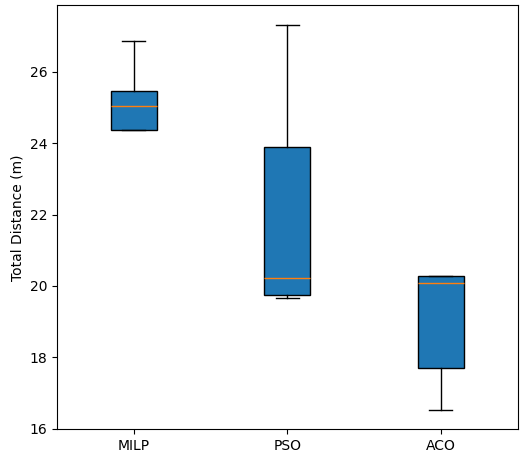}
        \caption{5 tasks}
        \label{fig:boxplot_5}
    \end{subfigure}
    \hfill
    \begin{subfigure}[b]{0.32\textwidth}
        \centering
        \includegraphics[width=\linewidth]{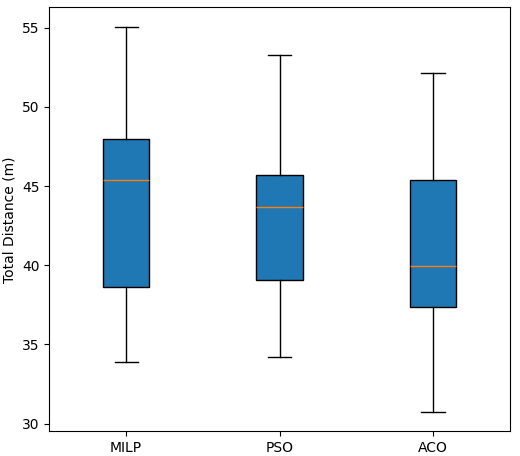}
        \caption{10 tasks}
        \label{fig:boxplot_10}
    \end{subfigure}
    \hfill
    \begin{subfigure}[b]{0.32\textwidth}
        \centering
        \includegraphics[width=\linewidth]{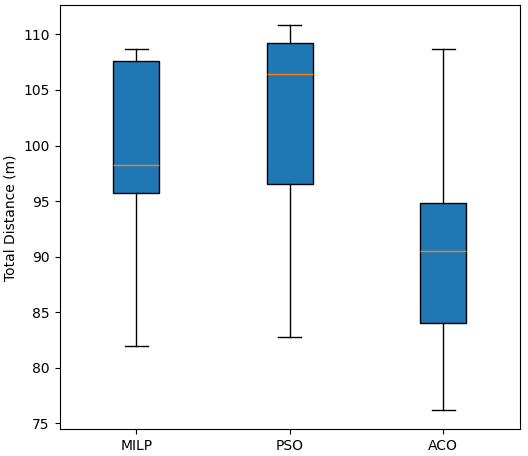}
        \caption{20 tasks}
        \label{fig:boxplot_20}
    \end{subfigure}
    \caption{Total travel distance distributions for ACO, PSO, and MILP across different task sizes.}
    \label{fig:boxplots_all}
\end{figure*}

This result is further confirmed in Figure \ref{fig:boxplots_all} that shows the total travel distance distributions achieved by ACO, PSO, and MILP across different task sizes. For the smallest task set (5 tasks), ACO consistently produces the lowest total distance with a narrow interquartile range, indicating both efficiency and stability. PSO shows slightly higher median performance but with large variability, suggesting inconsistent convergence, while MILP yields the highest and most uniform distances, reflecting deterministic yet less optimal solutions under increased constraints. As the number of tasks increases to 10 and 20, all methods exhibit greater dispersion due to the growing problem complexity. Nevertheless, ACO maintains a lower median distance than both PSO and MILP, particularly in the 20-task scenario, where MILP and PSO distances increase substantially. These results confirm that ACO scales more effectively and achieves a better balance between exploration and exploitation to produce shorter and more consistent routes.

\section{Conclusion}
In this work, we introduced a bi-layer ACO algorithm for MRTA that jointly optimizes assignment and sequencing decisions for delivery applications. In ROS 2/Gazebo simulations with three robots, the proposed method consistently outperforms PSO and MILP across all tested task sizes. It reduces total travel distance by up to 17.7\% compared with MILP and 9.8\% compared with PSO, and shortens completion time by up to 20\%. Lower standard deviations further indicate stable convergence. These results confirm that the proposed bi-layer approach enhances efficiency and scalability in multi-robot delivery scenarios. Future work will address dynamic environments, heterogeneous robot fleets, and communication constraints to validate the approach in real-world deployments.



\bibliographystyle{ACM-Reference-Format}
\bibliography{ref}

\end{document}